\documentclass[letterpaper]{article} 
\usepackage{aaai25}  
\usepackage{times}  
\usepackage{helvet}  
\usepackage{courier}  
\usepackage[hyphens]{url}  
\usepackage{graphicx} 
\usepackage{natbib}  
\usepackage{caption} 
\usepackage{algorithm}
\usepackage{algorithmic}
\usepackage{multirow}
\usepackage{makecell}
\usepackage{booktabs}
\usepackage{subcaption}
\usepackage{amssymb}
\usepackage{amsmath}
\usepackage{xcolor}

\usepackage{newfloat}
\usepackage{listings}
\DeclareCaptionStyle{ruled}{labelfont=normalfont,labelsep=colon,strut=off} 
\floatstyle{ruled}
\newfloat{listing}{tb}{lst}{}
\floatname{listing}{Listing}
\title{Interpretable Face Anti-Spoofing: Enhancing Generalization \\ with Multimodal Large Language Models}
\author{
    Guosheng Zhang\equalcontrib\textsuperscript{\rm1},
    Keyao Wang\equalcontrib\textsuperscript{\rm1},
    Haixiao Yue\equalcontrib\textsuperscript{\rm1},
    Ajian Liu\thanks{Corresponding author}\textsuperscript{\rm2},
    Gang Zhang\textsuperscript{\rm1}, \\
    Kun Yao\textsuperscript{\rm1},
    Errui Ding\textsuperscript{\rm1},
    Jingdong Wang\textsuperscript{\rm1}
}
\affiliations{
    \textsuperscript{\rm 1}Department of Computer Vision Technology (VIS), Baidu Inc \\
    \textsuperscript{\rm 2}CBSR\&MAIS, Institute of Automation, Chinese Academy of Sciences (CASIA) \\

    \{zhangguosheng, wangkeyao, yuehaixiao, zhanggang03, yaokun, dingerrui\}@baidu.com, 
    ajian.liu@ia.ac.cn, wangjingdong@outlook.com
}

\usepackage{bibentry}

\begin{document}

\maketitle

\begin{abstract}
Face Anti-Spoofing (FAS) is essential for ensuring the security and reliability of facial recognition systems. Most existing FAS methods are formulated as binary classification tasks, providing confidence scores without interpretation. They exhibit limited generalization in out-of-domain scenarios, such as new environments or unseen spoofing types. In this work, we introduce a multimodal large language model (MLLM) framework for FAS, termed Interpretable Face Anti-Spoofing (I-FAS), which transforms the FAS task into an interpretable visual question answering (VQA) paradigm. Specifically, we propose a Spoof-aware Captioning and Filtering (SCF) strategy to generate high-quality captions for FAS images, enriching the model's supervision with natural language interpretations. To mitigate the impact of noisy captions during training, we develop a Lopsided Language Model (L-LM) loss function that separates loss calculations for judgment and interpretation, prioritizing the optimization of the former. Furthermore, to enhance the model's perception of global visual features, we design a Globally Aware Connector (GAC) to align multi-level visual representations with the language model. Extensive experiments on standard and newly devised One to Eleven cross-domain benchmarks, comprising 12 public datasets, demonstrate that our method significantly outperforms state-of-the-art methods.

\end{abstract}

%

\begin{figure}[!ht]
    \centering
    \includegraphics[width=19.5pc]{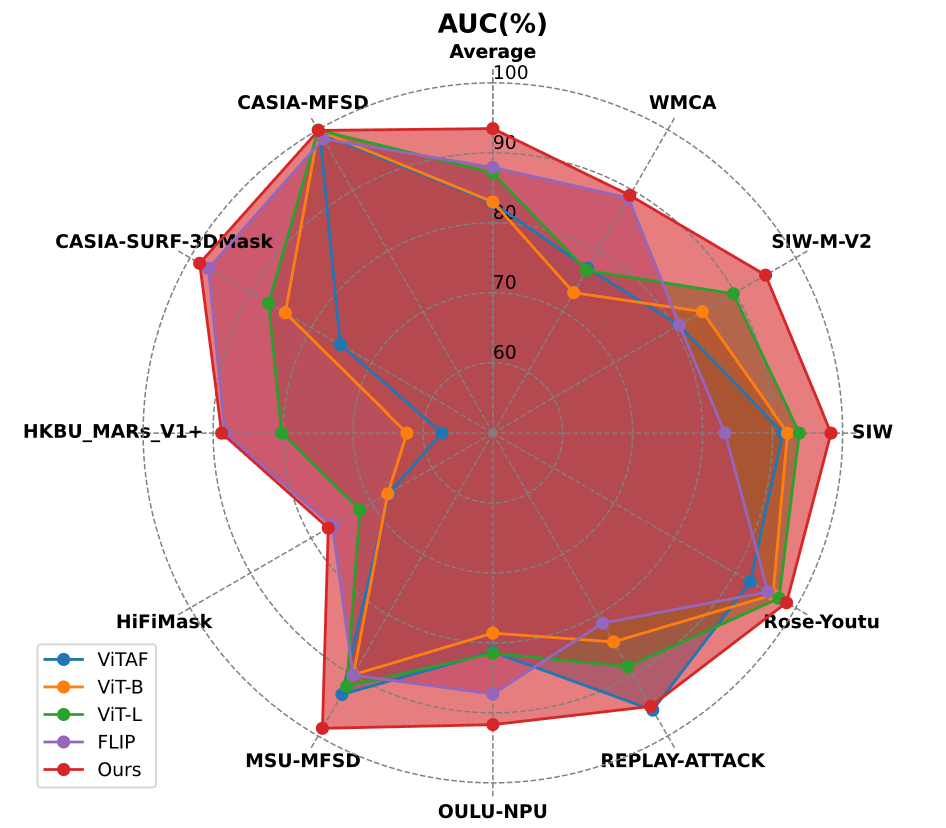}
    \caption{Comparison of performance under a challenging One to Eleven benchmark, where training is restricted to a single source domain (CelebA-Spoof) while testing across 11 target domains. The graph illustrates the notable superiority of our method (red point) compared to existing methods under the condition of a limited source domain.}
    \label{AUC}
\end{figure}

\section{Introduction}
Facial recognition technology has become increasingly sophisticated and is widely utilized for its inherent convenience and contactless operation. These systems are effectively employed in various applications, particularly in online payment and identity verification. However, they remain vulnerable to environmental fluctuations and diversified spoofing types, such as printed images, video replays~\cite{boulkenafet2017oulu}, and even high-fidelity 3D masks~\cite{liu2022contrastive}. Consequently, face anti-spoofing is developed to enhance the security and effectiveness of facial recognition systems in a variety of applications. Previous FAS methods~\cite{liu2018learning,yu2020searching} have demonstrated significant achievements, particularly in intra-domain scenarios, but they generally exhibit poor generalization when applied to cross-domain contexts, encountering novel environments, or spoof instruments.

One prevalent approach to improving cross-domain generalization is Domain Adaptation (DA)~\cite{wang2021self,liu2022source,zhou2022generative,yue2023cyclically,liu2024source}, which aims to reduce the discrepancy between the source and target domains. However, these methods often require access to target domain data during training, which is not always feasible. Another alternative approach is Domain Generalization (DG)~\cite{cai2022learning,zhou2023instance,cai2024towards,liu2024moeit,le2024gradient,zhou2024test}, designed to train models that can generalize robustly to the target domain by accessing multiple source domains. DG-based methods typically learn a theoretically domain-invariant feature representation across multiple source domains, employing techniques such as adversarial training or feature disentanglement. However, the substantial distributional differences between source domains, coupled with the practical challenges of diverse spoofing types, render the reliance on category-level annotations alone insufficient for the model to learn domain-invariant features.

Recent research \cite{srivatsan2023flip,liu2024cfpl,10.1145/3664647.3680856,guo2024style,shi2024shield} has proposed employing a CLIP-like framework for FAS. This approach outperforms the traditional unimodal approach by incorporating a textual modality that provides descriptive textual information for FAS images, thereby aiding in model training and decision-making. Although utilizing semantic content from the text significantly enhances the model's generalization capabilities, the reliance on manually constructed text based on prior knowledge leads to limited diversity and a scarcity of instructive information.

Rethinking how humans effortlessly and robustly identify spoofs, we observe that they can disregard irrelevant factors, concentrate on key spoofing cues, and derive interpretable judgments through causal reasoning. Inspired by recent breakthroughs in multimodal large language models (MLLMs), which have demonstrated exceptional image-text comprehension capabilities and strong generalization capabilities. We propose a pioneering MLLM framework for FAS, termed Interpretable FAS (I-FAS), which transforms the FAS classification task into an interpretable framework of Visual Question Answering (VQA), providing additional natural language interpretations for judgments. However, current FAS datasets typically rely on category-level annotations, lacking granular details such as comprehensive image descriptions, especially those that highlight spoofing clues. To bridge this gap, we design the Spoof-aware Captioning and Filtering (SCF) strategy. This strategy integrates two distinct captioners: a general captioner that generates captions for real samples, and a spoof-aware captioner that specializes in providing captions with spoof-specific preferences, such as the type, medium, and form. To mitigate the impact of noisy captions during training, we develop a Lopsided Language Model (L-LM) loss function that differentiates between the judgment and the interpretation components of the answer, allowing for separate loss calculations. By increasing the loss weight attributed to the judgment, we substantially accelerate and enhance the stability of model convergence. Furthermore, to enhance the FAS model's perception of global visual features, particularly focusing on low-level visual features like moiré patterns from screens and blur associated with paper, we introduce the Globally Aware Connector (GAC). The GAC integrates multi-level global representation, providing the language model with a comprehensive understanding from the global to the local perspectives. Finally, to better demonstrate the generalizability of our approach, we have established a more extensive and challenging benchmark, One to Eleven, including 12 public FAS datasets. Our method elucidates the rationale behind its decisions, thereby enhancing interpretability and robustness, and improving cross-domain generalization performance.
\begin{itemize}
    \item We reformulate the FAS classification task into an Interpretable VAQ paradigm, termed I-FAS, which leverages a Globally Aware Connector (GAC) to capture multi-level spoof cues, thereby enhancing robustness and cross-domain generalization.
    \item We introduce Spoof-aware Captioning and Filtering (SCF) to provide more comprehensive annotation data for FAS, enabling the model to explain its decision-making process, thereby improving interpretability.
    \item We developed a Lopsided LM (L-LM) loss that separates the answer into judgment and interpretation for separate loss computation, mitigating the impact of noisy captions during training.
    \item Extensive experiments on both standard and newly designed (One to Eleven) cross-domain benchmarks demonstrate that our method achieves a significant improvement over state-of-the-art methods.    
\end{itemize}

\begin{figure*}[tb]
    \centering
    \includegraphics[width=42pc]{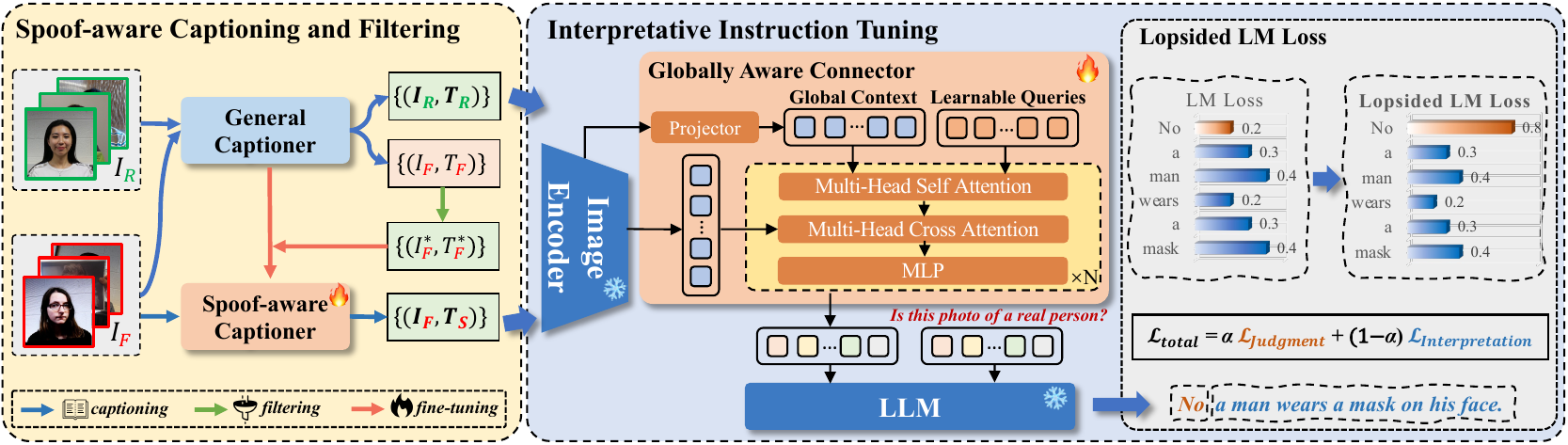}
    \caption{Overview of the proposed Interpretable Face Anti-Spoofing (I-FAS) framework. section illustrates the process of our proposed Spoof-aware Captioning and Filtering (SCF) strategy. The central section details the model architecture, which includes a frozen visual encoder, a pre-trained language model (LLM), and the Globally Aware Connector (GAC). The rightmost section presents a schematic representation of the Lopsided Language Model (L-LM) loss.}
    \label{intro}
\end{figure*}

\section{Related Work}
\subsection{Face Anti-Spoofing}
FAS aims to determine whether an image captures a genuine or a deceptive presentation attack. Early research was rooted in traditional handcrafted features and evolved into more advanced deep learning techniques\cite{wang2022patchnet,zhang2020face,wang2024csdg}. \cite{yu2020searching} have focused on designing specialized architectures for FAS, including the recently superior effective transformer structures highlighted by \cite{huang2022adaptive}. Furthermore, advancements have been facilitated through auxiliary supervision signals, such as depth maps \cite{liu2018learning} and reflection maps \cite{zhang2021structure}. Despite their success in intra-dataset scenarios, these methods often falter in cross-domain settings. To address this issue, recent methods have employed DA-based techniques \cite{wang2021self,liu2022source,zhou2022generative,yue2023cyclically,liu2024source} to reduce inter-domain distribution discrepancies by introducing unlabeled target domain data. Concurrently, DG-based approaches \cite{zhou2023instance,cai2024towards,liu2024moeit,le2024gradient,zhou2024test} aim to learn domain-invariant features across multiple source domains via adversarial learning \cite{shao2019multi,jia2020single}, meta-learning \cite{cai2022learning,chen2021generalizable}. Additionally, incremental learning (IL) methods \cite{hu2024domain,guo2022multi,cai2023rehearsal,wang2024multi} are considered to tackle the catastrophic forgetting problem in the context of domain discontinuity. However, these methods concentrate on extracting generalized liveness-specific features, relying exclusively on binary labels. 

\subsection{Vision-Language Models}
Recently, multimodal vision-language models have seen remarkable advancements, delivering promising performance across a spectrum of tasks. Pioneering work \cite{radford2021learning,huang2024empiricalstudyllama3quantization,jiang2024effectiveness} focused on vision-language pre-training, which aimed to cultivate foundational models capable of enhanced performance on diverse vision-language tasks. Recent methods \cite{liu2024visual,li2023blip} have harnessed knowledge from frozen LLMs \cite{zhang2022opt} for vision-to-language generation tasks in an instruction-tuning manner, commonly referred to as multimodal large language models (MLLMs). The pivotal challenge is the design of a cross-modal connector. This connector must align visual features from the pre-trained visual backbone to the word embedding space of the language model. Vision-Language models (VLM) have catalyzed innovative thinking and progress in numerous disciplines~\cite{yuan2024osprey,zanella2024harnessing,guo2023hierarchical,CLIP-forgery-detection}. In terms of FAS, \cite{srivatsan2023flip} have adapted the multimodal pre-trained CLIP model, grounding visual representations with natural language supervision to enhance generalizability. \cite{liu2024cfpl} have designed a CLIP-like model to expand the semantic space through textual prompt learning, thereby fine-tuning visual features for improved generalization. Instead of adopting inherent language templates like~\cite{deepfakeVQA}, our approach leverages interpretable and spoof-aware text to furnish the model with valuable supervisory signals, thereby enriching its learning process and decision-making capabilities.

\section{Methodology}
\subsection{Spoof-aware Captioning and Filtering}
\begin{figure}
    \centering
    \includegraphics[width=20pc]{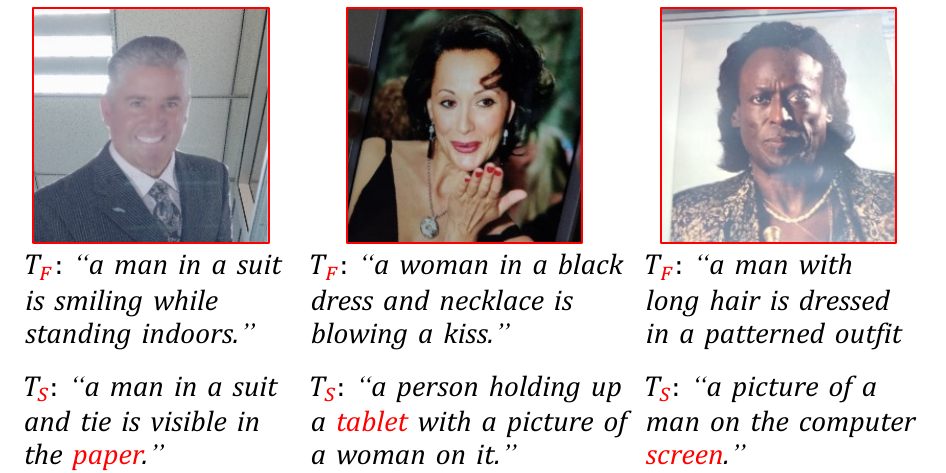}
    \caption{Illustration of some image-caption pair from the spoof sample. The captions $T_{F}$ and $T_{S}$ are generated by general captioner $C_{G}$ and spoof-aware captioner $C_{S}$, respectively. The keywords instrumental in identifying spoof cues are distinctly highlighted in red within the captions.}
    \label{scf_sample}
\end{figure}

Off-the-shelf captioners, trained on a broad range of generic scene images, often focus on general attributes like expressions, clothing, and environment when generating descriptions for FAS images, as shown in Figure~\ref{scf_sample}. They exhibit significant limitations in identifying specific spoof-related cues, such as the type, medium, and form. To address this, we develop a Spoof-aware Captioning and Filtering (SCF) strategy, which aims to generate interpretative descriptions elucidating the rationale for classifying images as either genuine or a presentation attack. As illustrated in Algorithm \ref{alg:algorithm}, we begin by aggregating 12 public FAS datasets into a cohesive dataset $\mathcal{D}=\{(I^{i}, Y^{i})\}_{i=1}^{N}$, where $I^{i}$ represents the images categorized as either real ($I_{R}$) or fake ($I_{F}$) and $Y^{i}$ denotes category label. Concurrently, we establish a spoof-aware keyword dictionary $\mathcal{K}$ categorized by spoof types, including print, replay, mask, and mannequin attacks. We initiate the process by generating captions for all fake images $I_{F}$ using an off-the-shelf general captioner $C_{G}$ \cite{li2023blip}, resulting in textual descriptions $T_{F}$. We then filter samples, retaining only those samples whose captions contain keywords from $\mathcal{K}$ that correspond to their spoof type, culminating in the formation of a spoof-aware dataset $\mathcal{D}_{S}$. For example, if a sample's caption contains the keyword \textit{``screen"} and is annotated as a screen attack, we can heuristically infer that the image likely contains discernible cues of the screen attack. Further detailed analysis can be found in the Appendix.

To ensure all fake samples are endowed with spoof-aware captions, we finetune the general captioner $C_{G}$ using our curated spoof-aware dataset $\mathcal{D}_{S}$, resulting in the spoof-aware captioner $C_{S}$, which we hypothesize is capable of inherently recognizing and describing spoof-related cues, as shown in Figure~\ref{scf_sample}. For real samples, we again employ $C_{G}$ to generate captions for all real samples $I_{R}$. Finally, by incorporating differential captions corresponding to the original category labels, we achieve a more comprehensive dataset that not only categorizes but also elucidates the rationale underpinning each judgment. 

\begin{algorithm}[tb]
\caption{Spoof-aware Captioning and Filtering}
\label{alg:algorithm}
\textbf{Input}: Dataset $\mathcal{D}=\{(I^{i}, Y^{i})\}_{i=1}^{N}$, where $I^{i} \in \{I_{R}, I_{F}\}$,
$Y^{i} \in \{Y_{R}, Y_{F}\}$, $F=\{print, replay, mask, mannequin\}$
Keywords: $\mathcal{K}=\{``paper": Y_{print}, ``screen": Y_{replay},...\}$ 
General captioner: $C_{G}$ \\
\textbf{Output}: Dataset $\mathcal{D}_{cap}=\{(I^{i},Y^{i}, T^{i})\}_{i=1}^{N}$,where $T^{i} \in \{T_{R}, T_{S}\}$ 
\begin{algorithmic}[1] 
\STATE Captioning $T_{F} = C_{G}(I_{F})$
\STATE Initialize empty dataset $\mathcal{D}_{S}$
\FOR{each sample $(I^{i}_{F},Y^{i}_{F}, T^{i}_{F})$}
\FOR{each keyword $k$ in $\mathcal{K}$}
\IF{$k$ in $T^{i}_{F}$ and $\mathcal{K}[k]$ match $Y^{i}_{F}$}
\STATE $\mathcal{D}_{S} \gets \mathcal{D}_{S} \cup \{(I^{i}_{F},Y^{i}_{F}, T^{i}_{F})\}$ 
\ENDIF
\ENDFOR
\ENDFOR
\STATE Finetune $C_{G}$ with $\mathcal{D}_{S}$ then obtain $C_{S}$
\STATE Captioning $T_{R} = C_{G}(I_{R})$ and $T_{S} = C_{S}(I_{F})$
\STATE $\mathcal{D}_{cap} \gets \{(I_{R},Y_{R}, T_{R})\} \cup \{(I_{F},Y_{F}, T_{S})\}$
\STATE \textbf{return} $\mathcal{D}_{cap}$
\end{algorithmic}
\end{algorithm}

\subsection{Interpretative Instruction Tuning}
\subsubsection{Revisiting MLLMs:}
MLLMs are designed to address sophisticated tasks by generating responses using multimodal inputs, including visual and textual data. The architecture of MLLMs comprises three principal components: 1) the pre-trained visual encoders $E_{V}$: This component converts the input image $I \in \mathbb{R}^{H \times W \times 3}$ into a set of visual features $X_{V} \in \mathbb{R}^{N \times D}$, where $N$ and $D$ denotes the count and dimension of visual features respectively. 2) Vision-Language Connector $P_{V \to T}$: This element is tasked with aligning visual features to textual token space $T$ of LLMs. 
3) LLMs $\Phi_{T}$: As the cornerstone of MLLMs, these models are capable of auto-regressively generating free-form responses when prompted with visual tokens $X_{V}$ and textual tokens $X_{T}$. Specifically, for a sequence of length $L$, the probability of generating target answers $X_{A}$ is computed by:
\begin{equation}
    p(X_{A}\mid X_{V},X_{T})=\prod_{i=1}^{L}p(X_{A,i}\mid X_{V},X_{T,<i},X_{A,<i})
\end{equation}
In the conventional MLLM framework, the effectiveness of the visual encoder and LLM is largely influenced by the pre-training phase. Consequently, the design of the connector is crucial during the instructing tuning stage, as it is responsible for the extraction and transformation of pertinent information from the visual features into the textual domain, effectively navigating through redundant visual information.

\subsubsection{Globally Aware Connector:}
The vision-language connector is designed to transform the visual features $X_{V}$ into textual tokens $X_{T}$ compatible with textual processing. To enhance the model’s perception of multi-level visual features, we introduce the Globally Aware Connector (GAC), which enhances global perception during visual feature extraction by incorporating multi-level global context features. As illustrated in Figure \ref{intro}, we utilize multi-layer global visual features $G_{V}=\{g_{1}, g_{2}, ..., g_{L}\}$, projected by linear layer, as additional input queries $Q_{V} \in \mathbb{R}^{L \times D}$, where $L$ is the total layers of the image encoder and $g_{i}$ represents \textit{cls} token of the \textit{i-th} layers. These queries engage with the learnable queries $Q_{P} \in \mathbb{R}^{M \times D}$ through multi-head self-attention layers (\texttt{MSA}), where $M$ denotes the number of queries. Concurrently, the learnable queries interact with local visual features $X_{V}$ via multi-head cross-attention operations (\texttt{MCA}). This process facilitates a comprehensive integration of global and local visual information and can be expressed as:
\begin{equation}
\begin{aligned}
        Q&=\texttt{Concat}(Q_{P}, Q_{V}) \\
        Q'&=Q+\texttt{MSA}(\texttt{LN}(Q)) \\
        Q''&=Q'+\texttt{MCA}(\texttt{LN}(Q'), \texttt{LN}(X_{V})) \\
        X_{T}&=Q'' + \texttt{MLP}(\texttt{LN}(Q''))
\end{aligned}
\end{equation}
Research \cite{jiang2023clip} indicates that different layers of visual encoder exhibit distinct biases towards various patterns: shallow layers are adept at capturing detailed low-level information, while deep layers are proficient in semantic comprehension. By this mechanism, the LLM receives visual information conducive to global perception. The GAC provides LLM with a comprehensive visual representation that integrates both the macroscopic context and the microscopic details. 

\subsubsection{Training with Lopsided LM Loss:}
In this work, we reformulate the FAS classification task into the VQA paradigm. For each training instance, we construct single-turn VQA data in the form of $(I, T_{Q}, T_{A})$, where $T_{Q}$ is a unified question: \textit{``Is this photo of a real person?"} serving as instruction. The target answers are formatted as follows:
\begin{equation}
    T_{A} = [T_{Judgment}, T_{Interpretation}]
\end{equation}
where $T_{Judgment} \in \{``Yes", ``No"\}$ indicates the judgment outcome, while $T_{Interpretation} = This\ is\ <T_{R}/T_{S}>$ comprises descriptive captions generated by the general captioner for real images ($T_{R}$) and the spoof-aware captioner for fake images ($T_{S}$).

As shown in Figure \ref{intro}, to mitigate the impact of noisy interpretations, we introduce a Lopsided Language Model (L-LM) Loss that separately calculates the loss for $T_{Judgment}$ and $T_{Interpretation}$, enabling the model to prioritize the accuracy of judgments. The training objective is defined as: 
\begin{equation}
    \mathcal{L}_{total} = \alpha \mathcal{L}_{Judgment} + (1 - \alpha) \mathcal{L}_{Interpretation}
\end{equation}
where $\alpha$ is a hyperparameter that balances the emphasis between judgment and interpretation loss components. In the inference phase, the model's prediction is determined by the probability assigned to the word \textit{``Yes"}, which indicates the likelihood of the sample being classified as real.

\begin{table*}[t]
    \small
    \centering
    \scalebox{0.92}{
    \begin{tabular}{p{40mm}<{\centering}p{10mm}<{\centering}p{10mm}<{\centering}p{10mm}<{\centering}p{10mm}<{\centering}p{10mm}<{\centering}p{10mm}<{\centering}p{10mm}<{\centering}p{10mm}<{\centering}p{10mm}<{\centering}}
    \toprule
    \multirow{2}{*}{Methods} & \multicolumn{2}{c}{O\&C\&I to M}   & \multicolumn{2}{c}{O\&M\&I to C} & \multicolumn{2}{c}{O\&C\&M to I} & \multicolumn{2}{c}{I\&C\&M to O} & Avg. \\ 
    \cmidrule(r){2-3}
    \cmidrule(r){4-5}
    \cmidrule(r){6-7}
    \cmidrule(r){8-9}
    \cmidrule(r){10-10}
    {}   & \multicolumn{1}{l}{HTER(\%)} & \multicolumn{1}{l}{AUC(\%)} & \multicolumn{1}{l}{HTER(\%)} & \multicolumn{1}{l}{AUC(\%)} & \multicolumn{1}{l}{HTER(\%)} & \multicolumn{1}{l}{AUC(\%)} & \multicolumn{1}{l}{HTER(\%)} & \multicolumn{1}{l}{AUC(\%)} & HTER(\%) \\ 
    \midrule
    FGHV \cite{liu2022feature} & 9.17 & 96.92 & 12.47 & 93.47 & 16.29 & 90.11 & 13.58 & 93.55 & 12.88 \\
    GDA \cite{zhou2022generative} & 9.20 & 98.00 & 12.20 & 93.00 & 10.00 & 96.00 & 14.40 & 92.60 & 11.45 \\
    PatchNet \cite{wang2022patchnet} & 7.10 & 98.46 & 11.33 & 94.58 & 13.40 & 95.67 & 11.82 & 95.07 & 10.91 \\
    SSAN \cite{wang2022domain} & 6.67 & 98.75 & 10.00 & 96.67 & 8.88 & 96.79 & 13.72 & 93.63 & 9.82 \\
    IADG \cite{zhou2023instance} & 5.41 & 98.19 & 8.70 & 96.40 & 10.62 & 94.50 & 8.86 & 97.14 & 8.40 \\
    UDG-FAS \cite{liu2023towards} & 5.95 & 98.47 & 9.82 & 96.76 & 5.86 & 98.62 & 10.97 & 95.36 & 8.15 \\
    TTDG \cite{zhou2024test} & 4.16 & 98.48 & 7.59 & 98.18 & 9.62 & 98.18 & 10.00 & 96.15 & 7.84 \\
    SA-FAS \cite{sun2023rethinking} & 5.95 & 96.55 & 8.78 & 95.37 & 6.58 & 97.54 & 10.00 & 96.23 & 7.83 \\
    DiVT-M \cite{liao2023domain} & 2.86 & 99.14 & 8.67 & 96.92 & 3.71 & \textbf{99.29} & 13.06 & 94.04 & 7.08 \\
    GAC-FAS \cite{le2024gradient} & 5.00 & 97.56 & 8.20 & 95.16 & 4.29 & 98.87 & 8.60 & 97.16 & 6.52 \\
    \midrule
    FLIP (Srivatsan et al. 2023) & 4.95 & 98.11 & 0.54 & 99.98 & 4.25 & 99.07 & 2.31 & 99.63 & 3.01 \\
    CFPL \cite{liu2024cfpl} & 1.43 & 99.28 & 2.56 & 99.10 & 5.43 & 98.41 & 2.50 & 99.42 & 2.98 \\
    Ours & \textbf{0.32} & \textbf{99.88} & \textbf{0.04} & \textbf{99.99} & \textbf{3.22} & 98.48 & \textbf{1.74} & \textbf{99.66} & \textbf{1.33} \\ 
    \bottomrule
    \end{tabular}}
    \caption{Comparison with the closest and SOTA FAS methods in Protocol 1 on MSU-MFSD (M), CASIA-FASD (C), ReplayAttack (I), and OULU-NPU (O) datasets. Avg indicates the average performance across four experimental scenarios. The scores presented in bold represent the best performance.}
    \label{prot1}
\end{table*}

\section{Experiments}
\subsection{Experimental Setup}
\subsubsection{Databases, Protocols, and Evaluation Metrics:}
We evaluate our method on two protocols. For Protocol 1, Following established practices, we implement the leave-one-domain-out testing approach on several datasets: MSU-MFSD (M)\cite{wen2015face}, CASIA-MFSD (C) \cite{zhang2012face}, Idiap Replay Attack (I) \cite{chingovska2012effectiveness}, and OULU-NPU (O) \cite{boulkenafet2017oulu}. To assess the robustness of our method in more demanding conditions, we set up Protocol 2 as One to Eleve testing protocol. Employing only CelebA-Spoof \cite{zhang2020celeba} as the source domain, and 11 datasets as target domains for cross-domain testing. This selection include MSU-MFSD \cite{wen2015face}, CASIA-MFSD \cite{zhang2012face}, Idiap Replay Attack \cite{chingovska2012effectiveness}, OULU-NPU \cite{boulkenafet2017oulu}, SIW \cite{liu2018learning}, Rose-Youtu \cite{li2018unsupervised}, HKBU-MARs-V1+ \cite{liu2018remote}, WMCA \cite{george2019biometric}, SIW-M-V2 \cite{guo2022multi}, CASIA-SURF-3DMask \cite{yu2020fas} and HiFiMask \cite{liu2022contrastive}. For both protocols, we utilize the Area Under the Curve (AUC) and the Half Total Error Rate (HTER) as our primary evaluation metrics. The higher AUC and lower HTER signify superior performance.

\subsubsection{Implementation Details:}
We crop the face images and resize them to $224 \times 224 \times 3$ with RGB channels. For the frozen image encoder, we utilize pre-trained vision models: ViT-L/14 from CLIP \cite{radford2021learning}. Following \cite{li2023blip}, OPT-2.7B \cite{zhang2022opt} is adopted as the pre-trained large language model. We use the AdamW optimizer, with an initial learning rate set to $10^{-5}$ and a weight decay parameter set to $10^{-2}$. We configure our training process with a batch size of 32 and a maximum of 10 epochs for both Protocol 1 and Protocol 2. For Protocol 2, we meticulously reproduce the baseline methods, including FLIP \cite{srivatsan2023flip} and ViTAF \cite{huang2022adaptive}, using the official code provided. Both ViT-B and ViT-L are pre-trained CLIP \cite{radford2021learning}. To ensure the integrity and reproducibility of our experiments, we report all results as the mean of three independent runs, each with a unique initialization seed.

\begin{table*}[t]
\small
    \centering
    \begin{subtable}[t]{0.23\linewidth}
    \caption{\textbf{Average Over 11 Datasets}}
    \scalebox{0.94}{
    \begin{tabular}{p{10mm}<{\centering}p{10mm}<{\centering}p{10mm}<{\centering}}
    \toprule
    Methods & HTER(\%) & AUC(\%) \\
    \midrule
    ViTAF & 23.85 & 82.82  \\
    ViT-B & 23.48 & 82.98  \\
    ViT-L & 21.08 & 85.61  \\
    FLIP & 18.73 & 87.90  \\
    Ours & \textbf{11.30} & \textbf{93.71} \\
    \bottomrule
    \end{tabular}}
    \end{subtable}
    ~
    \centering
    \begin{subtable}[t]{0.23\linewidth}
    \caption{\textbf{CASIA-MFSD}}
    \scalebox{0.94}{
    \begin{tabular}{p{10mm}<{\centering}p{10mm}<{\centering}p{10mm}<{\centering}}
    \toprule
    Methods & HTER(\%) & AUC(\%) \\
    \midrule
    ViTAF & 3.11 & 99.48  \\
    ViT-B & \textbf{0.70} & 99.86  \\
    ViT-L & 0.93 & \textbf{99.95}  \\
    FLIP & 4.88 & 98.48  \\
    Ours & 1.11 & 99.88 \\
    \bottomrule
    \end{tabular}}
    \end{subtable}
    ~
    \centering
    \begin{subtable}[t]{0.23\linewidth}
    \caption{\textbf{CASIA-SURF-3DMask}}
    \scalebox{0.94}{
    \begin{tabular}{p{10mm}<{\centering}p{10mm}<{\centering}p{10mm}<{\centering}}
    \toprule
    Methods & HTER(\%) & AUC(\%) \\
    \midrule
    ViTAF & 32.44 & 75.20  \\
    ViT-B & 24.89 & 84.26  \\
    ViT-L & 23.54 & 84.22  \\
    FLIP & 8.83 & 96.93  \\
    Ours & \textbf{6.18} & \textbf{98.40} \\
    \bottomrule
    \end{tabular}}
    \end{subtable}
    ~
    \centering
    \begin{subtable}[t]{0.23\linewidth}
    \caption{\textbf{HKBU-MARs-V1+}}
    \scalebox{0.94}{
    \begin{tabular}{p{10mm}<{\centering}p{10mm}<{\centering}p{10mm}<{\centering}}
    \toprule
    Methods & HTER(\%) & AUC(\%) \\
    \midrule
    ViTAF & 49.29 & 57.28  \\
    ViT-B & 45.08 & 62.28  \\
    ViT-L & 33.33 & 73.88  \\
    FLIP & \textbf{17.25} & 88.31  \\
    Ours & 18.64 & \textbf{88.77} \\
    \bottomrule
    \end{tabular}}
    \end{subtable}
    ~
    \centering
    \begin{subtable}[t]{0.23\linewidth}
    \caption{\textbf{HiFiMask}}
    \scalebox{0.94}{
    \begin{tabular}{p{10mm}<{\centering}p{10mm}<{\centering}p{10mm}<{\centering}}
    \toprule
    Methods & HTER(\%) & AUC(\%) \\
    \midrule
    ViTAF & 37.30 & 67.10  \\
    ViT-B & 37.33 & 67.35  \\
    ViT-L & 32.81 & 72.58  \\
    FLIP & 28.32 & 76.50  \\
    Ours & \textbf{28.23} & \textbf{77.17} \\
    \bottomrule
    \end{tabular}}
    \end{subtable}
    ~
    \centering
    \begin{subtable}[t]{0.23\linewidth}
    \caption{\textbf{MSU-MFSD}}
    \scalebox{0.94}{
    \begin{tabular}{p{10mm}<{\centering}p{10mm}<{\centering}p{10mm}<{\centering}}
    \toprule
    Methods & HTER(\%) & AUC(\%) \\
    \midrule
    ViTAF & 12.86 & 93.14  \\
    ViT-B & 16.67 & 89.89  \\
    ViT-L & 20.87 & 85.65  \\
    FLIP & 19.37 & 89.95  \\
    Ours & \textbf{5.63} & \textbf{98.73} \\
    \bottomrule
    \end{tabular}}
    \end{subtable}
    ~
    \centering
    \begin{subtable}[t]{0.23\linewidth}
    \caption{\textbf{OULU-NPU}}
    \scalebox{0.94}{
    \begin{tabular}{p{10mm}<{\centering}p{10mm}<{\centering}p{10mm}<{\centering}}
    \toprule
    Methods & HTER(\%) & AUC(\%) \\
    \midrule
    ViTAF & 26.73 & 81.28  \\
    ViT-B & 28.53 & 78.59  \\
    ViT-L & 29.42 & 78.07  \\
    FLIP & 20.57 & 87.30  \\
    Ours & \textbf{14.86} & \textbf{91.68} \\
    \bottomrule
    \end{tabular}}
    \end{subtable}
    ~
    \centering
    \begin{subtable}[t]{0.23\linewidth}
    \caption{\textbf{REPLAY-ATTACK}}
    \scalebox{0.94}{
    \begin{tabular}{p{10mm}<{\centering}p{10mm}<{\centering}p{10mm}<{\centering}}
    \toprule
    Methods & HTER(\%) & AUC(\%) \\
    \midrule
    ViTAF & 12.38 & \textbf{95.73}  \\
    ViT-B & 24.80 & 84.47  \\
    ViT-L & 16.58 & 92.00  \\
    FLIP & 25.67 & 81.37  \\
    Ours & \textbf{9.15} & 95.12 \\
    \bottomrule
    \end{tabular}}
    \end{subtable}
    ~
    \centering
    \begin{subtable}[t]{0.23\linewidth}
    \caption{\textbf{Rose-Youtu}}
    \scalebox{0.94}{
    \begin{tabular}{p{10mm}<{\centering}p{10mm}<{\centering}p{10mm}<{\centering}}
    \toprule
    Methods & HTER(\%) & AUC(\%) \\
    \midrule
    ViTAF & 69.34 & 74.22  \\
    ViT-B & 82.69 & 63.22  \\
    ViT-L & 80.47 & 71.69  \\
    FLIP & 80.73 & 73.60  \\
    Ours & \textbf{5.52} & \textbf{98.48} \\
    \bottomrule
    \end{tabular}}
    \end{subtable}
    ~
    \centering
    \begin{subtable}[t]{0.23\linewidth}
    \caption{\textbf{SIW}}
    \scalebox{0.94}{
    \begin{tabular}{p{10mm}<{\centering}p{10mm}<{\centering}p{10mm}<{\centering}}
    \toprule
    Methods & HTER(\%) & AUC(\%) \\
    \midrule
    ViTAF & 14.74 & 92.51  \\
    ViT-B & 9.13 & 96.24  \\
    ViT-L & 9.03 & 96.56  \\
    FLIP & 11.01 & 95.40  \\
    Ours & \textbf{4.02} & \textbf{98.34} \\
    \bottomrule
    \end{tabular}}
    \end{subtable}
    ~
    \centering
    \begin{subtable}[t]{0.23\linewidth}
    \caption{\textbf{SIW-M-V2}}
    \scalebox{0.94}{
    \begin{tabular}{p{10mm}<{\centering}p{10mm}<{\centering}p{10mm}<{\centering}}
    \toprule
    Methods & HTER(\%) & AUC(\%) \\
    \midrule
    ViTAF & 26.72 & 80.70  \\
    ViT-B & 22.60 & 84.59  \\
    ViT-L & 17.26 & 90.37  \\
    FLIP & 25.95 & 80.78  \\
    Ours & \textbf{10.89} & \textbf{95.02} \\
    \bottomrule
    \end{tabular}}
    \end{subtable}
    ~
    \centering
    \begin{subtable}[t]{0.23\linewidth}
    \caption{\textbf{WMCA}}
    \scalebox{0.94}{
    \begin{tabular}{p{10mm}<{\centering}p{10mm}<{\centering}p{10mm}<{\centering}}
    \toprule
    Methods & HTER(\%) & AUC(\%) \\
    \midrule
    ViTAF & 29.88 & 77.14  \\
    ViT-B & 34.72 & 73.10  \\
    ViT-L & 34.39 & 75.13  \\
    FLIP & \textbf{19.36} & 88.73  \\
    Ours & 20.07 & \textbf{89.17} \\
    \bottomrule
    \end{tabular}}
    \end{subtable}
    \caption{Comparison in Protocol 2, illustrating the challenge of training solely on the CelebA-Spoof dataset followed by testing across 11 distinct datasets. We run each experiment 3 times under different seeds and report the average HTER and AUC.}
    \label{prot2}
\end{table*}

\subsection{Comparison Results}
\subsubsection{Comparison Results in Protocol 1:}
 We evaluate our method using four leave-one-domain-out settings and compare its performance with that of the latest state-of-the-art approaches. As shown in Table \ref{prot1}, our method demonstrates a significant performance advantage over all single-modal methods, outperforming them by a substantial margin. This substantial advantage highlights the efficacy of multimodal learning in enhancing the model's generalizability. Furthermore, our approach also surpasses recent multimodal methods, such as \cite{srivatsan2023flip} and \cite{liu2024cfpl}, as evidenced by the average HTER reductions to 1.33\% from 2.98\% and 3.01\%. This improvement suggests that the incorporation of richer and more insightful textual information facilitates the acquisition of more robust and widely applicable features by the model.

\subsubsection{Comparison Results in Protocol 2:}
To further evaluate the generalizability of our approach, we utilize a single dataset as the source domain and perform cross-domain testing across 11 distinct datasets. As shown in Table~\ref{prot2}, our method demonstrates a compelling advantage of more than 7\% relative to the current state-of-the-art techniques. This substantial advantage is particularly evident under severely constrained source domain conditions. This scenario reflects real-world challenges that require robustness against emerging attack vectors and shifts in environmental conditions. Significantly, the datasets CASIA-SURF-3DMask, HKBU-MARs-V1+, and SIW-M-V2 encompass attack modalities absent in the source domain (CelebA-Spoof), including sophisticated 3D attacks, makeup alterations, and novel material-based attacks. Under these challenging conditions, our method consistently outperforms conventional single-modal DG models, demonstrating the model's superior adaptability and resilience.

\begin{figure*}
    \centering
    \includegraphics[width=42pc]{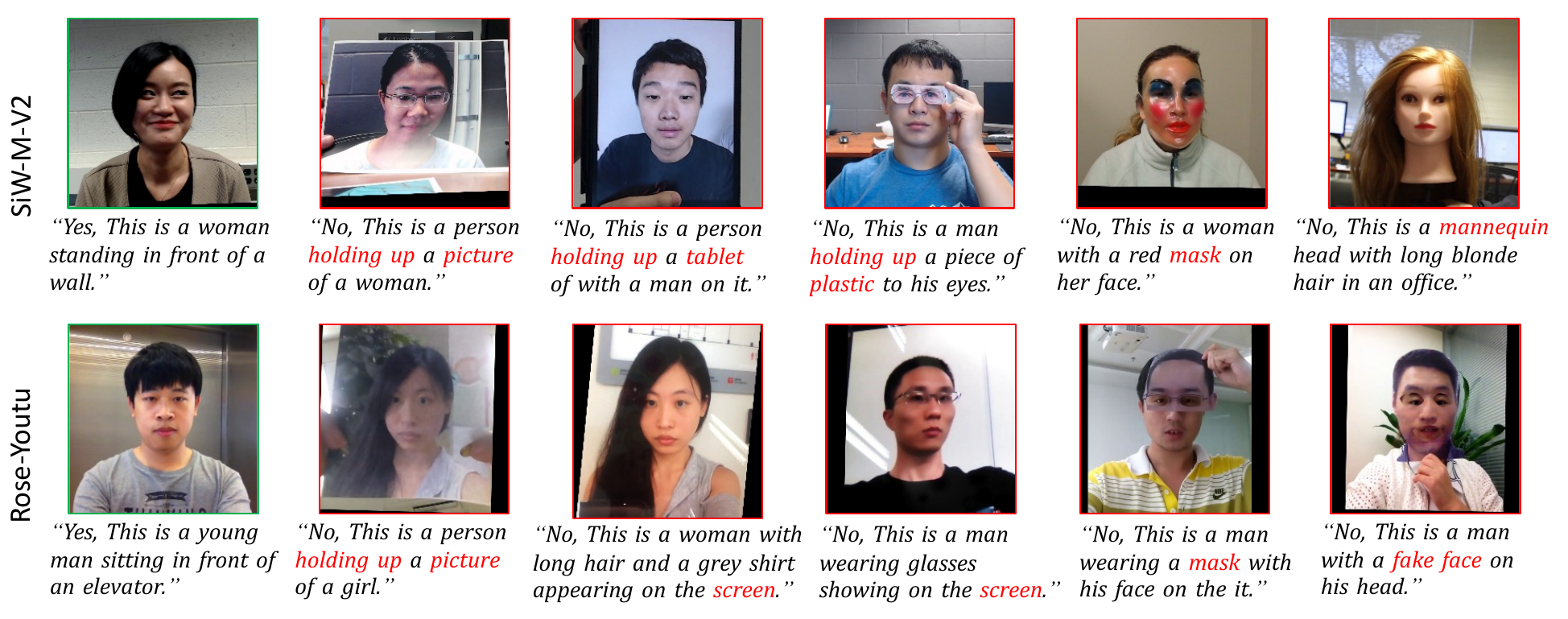}
    \caption{The output responses of I-FAS on some images from SIW-M-V2 and Rose-Youtu dataset under the unified question: \textit{``Is this photo of a real person?"}. The green box represents the real person, and the red box represents the spoof sample.}
    \label{answer}
\end{figure*}

\begin{figure}
    \centering
    \includegraphics[width=20pc]{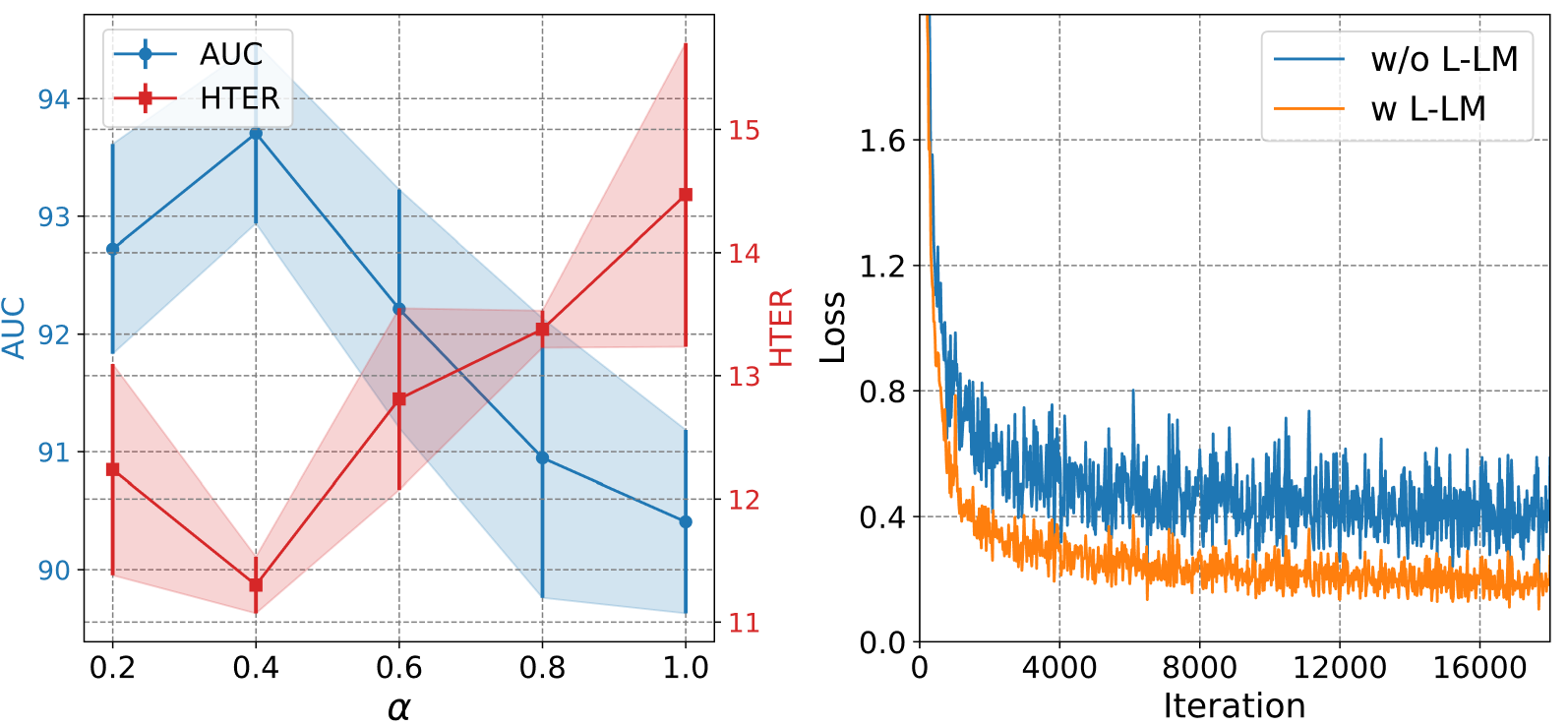}
    \caption{Left: Ablation analysis of hyperparameters $\alpha$ of lopsided LM loss in Protocol 2. Right: Visualization of loss convergence behavior with and without lopsided LM loss.}
    \label{aba}
\end{figure}

\begin{table}[t]
    \small
    \centering
    \scalebox{0.95}{
    \begin{tabular}{p{15mm}<{\centering}p{15mm}<{\centering}p{15mm}<{\centering}p{10mm}<{\centering}p{10mm}<{\centering}}
    \toprule
    SCF & GAC & L-LM & HTER(\%) & AUC(\%) \\
    \midrule
               & \checkmark & \checkmark & 14.47 & 90.41  \\
    \checkmark &   & \checkmark & 14.84 & 90.78  \\
    \checkmark & \checkmark &   & 12.06 & 92.81  \\
    \checkmark & \checkmark & \checkmark & \textbf{11.30} & \textbf{93.71} \\
    \bottomrule
    \end{tabular}}
    \caption{Ablation study on the effectiveness of key components within our proposed method, including the Spoof-aware Captioning and Filtering (SCF), Globally Aware Connector (GAC), and Lopsided LM (L-LM) Loss. The results are the average HTER and AUC in Protocol 2.}
    \label{table3}
\end{table}

\begin{table}[t]
    \small
    \centering
    \scalebox{0.95}{
    \begin{tabular}{p{10mm}<{\centering}p{40mm}<{\centering}p{10mm}<{\centering}p{10mm}<{\centering}}
    \toprule
    Methods & Text Format & HTER(\%) & AUC(\%) \\
    \midrule
    FLIP & $Template$ & 18.73 & 87.90 \\
    FLIP & $T_{Judgment}, \{T_{R}, T_{S}\}$ & 17.30 & 89.15 \\
    \midrule
    Ours & $T_{Judgment}$ & 14.47 & 90.41  \\
    Ours & $T_{Judgment}, \{T_{R}, T_{F}\}$ & 14.23 & 90.22 \\
    Ours & $T_{Judgment}, \{T_{R}, T_{S}\}$ & \textbf{11.30} & \textbf{93.71} \\
    \bottomrule
    \end{tabular}}
    \caption{Ablation study on the effects of various text formats as supervision. The results are the average HTER and AUC in Protocol 2.}
    \label{table4}
\end{table}

\subsection{Ablation Study}
\subsubsection{Study on Each Component.}
To systematically assess the impact of individual components within our framework, we perform an ablation study of the SCF, the GAC, and the L-LM Loss. For each component, we conduct sequential removal experiments within the framework and report the average results for HTER and AUC in Protocol 2.

As shown in Table 3, the first row corresponds to the scenario where the SCF strategy was removed, replacing the interpretable target answer with a simple binary judgment (\textit{``Yes/No"}). This modification resulted in a significant performance degradation, with an increase of 3.17\% in HTER and a decrease of 3.30\% in AUC. The second row indicates the impact of excluding the GAC module. To maintain experimental integrity and a consistent token number for LLMs, we replaced the multi-level global features with learnable queries. This alteration also culminated in a marked performance drop of +3.54\% (HTER) and -2.93\% (AUC), respectively. The third row presents the results of employing a standard LM Loss in place of our proposed Lopsided LM Loss. This change likewise resulted in a minor yet discernible reduction in the model's generalization capability, as indicated by a slight performance drop.

\subsubsection{Study on Spoof-aware Interpretation.}
To highlight the advantages of interpretable annotations, we apply them to the existing multimodal FAS method FLIP \cite{srivatsan2023flip}. Specifically, we only replace the original manual text template with spoof-aware interpretations. The Table \ref{table4} show a significant improvement with a 1.43\% decrease in HTER and a 1.25\% increase in AUC. To isolate the impact of caption diversity from the interpretability of spoof clues, we further replace the spoof-aware captions $T_{S}$ with those ($T_{F}$) generated by a general captioner devoid of spoof-specific preferences. Comparing the results from the second to last row and the third to last row in Table 4, it's clear that diverse captions alone do not enhance FAS judgments. In contrast, our spoof-aware interpretations are demonstrated to confer substantial improvements in the model's generalization capabilities.

\subsubsection{Study on Lopsided LM Loss.}
Figure \ref{aba} (Left) presents an ablation analysis of the hyperparameter $\alpha$ in lopsided LM loss, which is used to balance the emphasis between judgment and interpretation loss components. the average and variance of the AUC and HTER in Protocol 2 across a range of $\alpha$ values. Our findings indicate that extreme values of $\alpha$ lead to performance degradation. We suggest that an overly large $\alpha$ may diminish the generalization benefits conferred by interpretation components, as it disproportionately emphasizes the judgment aspect. Conversely, placing excessive weight on interpretation may introduce noise, thereby hindering the model's ability to learn effectively. As shown in Figure \ref{aba} (Right), not appropriately increasing the importance of judgment through the lopsided LM loss can result in slow and unstable model convergence during training.

\subsection{Visualization and Analysis}
Figure \ref{scf_sample} displays a selection of image-text pairs from the CelebA-Spoof dataset, crafted by our SCF strategy. Captions $T_{F}$ generated by the general captioner $C_{G}$ offer generic descriptions, capturing general attributes without bias toward specific features. In contrast, captions $T_{S}$ crafted by the spoof-aware captioner $C_{S}$ reveal the underlying spoofing tactics by emphasizing keywords such as \textit{``paper"}, \textit{``tablet"}, and \textit{``screen"}. Such annotations provide invaluable supervisory signals that are instrumental in the learning process of FAS tasks, enabling the model to discern subtle yet critical cues indicative of spoofing attempts.

In Protocol 2, after training solely on the CelebA-Spoof dataset, our model is tested on multiple unseen datasets. Figure \ref{answer} illustrates the model's output responses post-interpretative instruction tuning. Our method excels not only in rendering accurate judgments but also in furnishing rationales for these decisions, especially for samples with attack behaviors. The dual capability of accurate classification and interpretability strengthens the robustness and generalizability of our approach, which is pivotal for real-world applications where understanding the rationale for the decisions is as crucial as the decisions themselves.

\section{Conclusion}
In this work, we introduce a novel perspective for FAS by transforming it into an interpretable VQA task. Incorporating SCF strategies allows the model to generate interpretive captions, significantly enhancing cross-domain generalization capabilities. Our proposed Lopsided Language Model Loss optimizes the training process. Furthermore, the GAC enhances the model's perception of multi-level global context features, providing substantial performance improvements for liveness-specific feature learning. Experiments on Protocols 1 and 2 demonstrate the exceptional performance of our method across multiple public FAS datasets. 


\appendix
\counterwithin{table}{section}
\counterwithin{figure}{section}
\counterwithin{equation}{section}
\clearpage
\section{Spoof-aware Captioning and Filtering}
\subsection{Dataset}
In Table \ref{dataset}, we detail the publicly FAS datasets leveraged in our experimental framework. We perform a unified preprocessing pipeline on these 12 FAS datasets, including video de-framing (retaining one frame every five), face detection, and alignment. This systematic process resulted in a dataset of 2.2 million face images, categorized into 0.7 million real samples and 1.5 million fake samples. We categorize the spoofing types into four distinct groups based on their inherent characteristics: print, replay, mask, and mannequin. Notably, We separate mannequin attacks as an independent category, while other mask types are aggregated under a unified mask category. This latter group encompasses a variety of mask forms, including makeup, partial, and masks constructed from diverse materials such as transparent, plaster, and silicone. Figure \ref{statis} (Left) shows the distribution statistics of the four spoofing types.

\begin{table}[!ht]
    \small
    \centering
    \scalebox{0.95}{
    \begin{tabular}{p{20mm}<{\centering}p{55mm}<{\centering}}
    \toprule
    Spoofing Types & Keywords \\
    \midrule
    \multirow{2}{*}{Print} & \textit{``paper", ``cardboard", ``paper card", ``poster", ``picture"} \\ 
    \multirow{2}{*}{Replay} & \textit{``screen", ``monitor", ``cell", ``phone", ``tablet", ``laptop", ``ipad", } \\
    \multirow{2}{*}{Mask} & \textit{``mask", ``sticker", ``plastic", ``fake face", ``plaster"} \\
    \multirow{2}{*}{Mannequin} & \textit{``mannequin", ``doll", ``statue", ``sculpture", ``fake head"} \\
    \bottomrule
    \end{tabular}}
    \caption{Spoof-aware keyword dictionary $\mathcal{K}$ established in SCF is categorized by spoofing type, including print, replay, mask, and mannequin attacks.}
    \label{keyword}
\end{table}

\begin{table*}[!t]
    \small
    \centering
    \scalebox{0.95}{
    \begin{tabular}{p{68mm}<{\centering}p{5mm}<{\centering}p{20mm}<{\centering}p{6mm}<{\centering}p{60mm}<{\centering}}
    \toprule
    Dataset & Year & Real/Fake & Sub. & Spoofing Types \\
    \midrule
    CASIA-MFSD \cite{zhang2012face} & 2012 & 150/450(V) & 50 & Print, Replay \\
    Replay Attack \cite{chingovska2012effectiveness} & 2012 & 200/1000(V) & 50 & Print, Replay \\
    MSU-MFSD \cite{wen2015face} & 2014 & 70/210(V) & 35 & Print, Replay \\
    OULU-NPU \cite{boulkenafet2017oulu} & 2017 & 720/2880(V) & 55 & Print, Replay \\
    SIW \cite{liu2018learning} & 2018 & 1320/3300(V) & 165 & Print, Replay \\
    Rose-Youtu \cite{li2018unsupervised} & 2018 & 500/2850(V) & 20 & Print, Replay, Mask(paper, crop-paper) \\
    HKBU-MARs-V1+ \cite{liu2018remote} & 2018 & 110/60(V) & 12 & Mask(hard resin) \\
    \multirow{2}{*}{WMCA \cite{george2019biometric}} & \multirow{2}{*}{2019} & \multirow{2}{*}{347/1332(V)} & \multirow{2}{*}{72} & Print, Replay, Partial(glasses), Mask(plastic, silicone, paper, Mannequin) \\
    \multirow{4}{*}{SIW-M-V2 \cite{guo2022multi}} & \multirow{4}{*}{2019} & \multirow{4}{*}{785/915(V)} & \multirow{4}{*}{493} & Print, Replay, Mask(hard resin, plastic, silicone, paper, Mannequin), Makeup(cosmetics, impersonation, Obfuscation), Partial(glasses, cut paper) \\
    CASIA-SURF-3DMask \cite{yu2020fas} & 2020 & 288/864(V) & 48 & Mask(mannequin) \\
    CelebA-Spoof \cite{zhang2020celeba} & 2020 & 156384/469153(I) & 10177 & Print, Replay, Mask(paper) \\
    HiFiMask \cite{liu2022contrastive} & 2021 & 13650/40950(V) & 75 & Mask(transparent, plaster, resin) \\
    \bottomrule
    \end{tabular}}
    \caption{The summary of FAS dataset we utilize in our experiments. We summarize the dataset year, number of samples, number of subjects, and spoofing types, where `I' and `V' denotes `images' and `videos', respectively, and `Sub.' is short for Subjects.}
    \label{dataset}
\end{table*}

\subsection{Filtering}
Table \ref{keyword} presents the details of keyword dictionary $\mathcal{K}$ defined in SCF strategy. Based on prior knowledge, we have collected a set of keywords indicative of specific spoof-related cues for spoofing types, including print, replay, mask, and mannequin. This spoof-aware dictionary is a critical tool for enhancing the filtering process, ensuring that the captions not only describe the scene but also highlight the presence of potential spoofing attempts. For example, the keyword \textit{``screen"} is associated with revealing replay attacks, where an image or video is played back on a screen to deceive the facial recognition system.

\begin{figure}[!t]
    \centering
    \includegraphics[width=20pc]{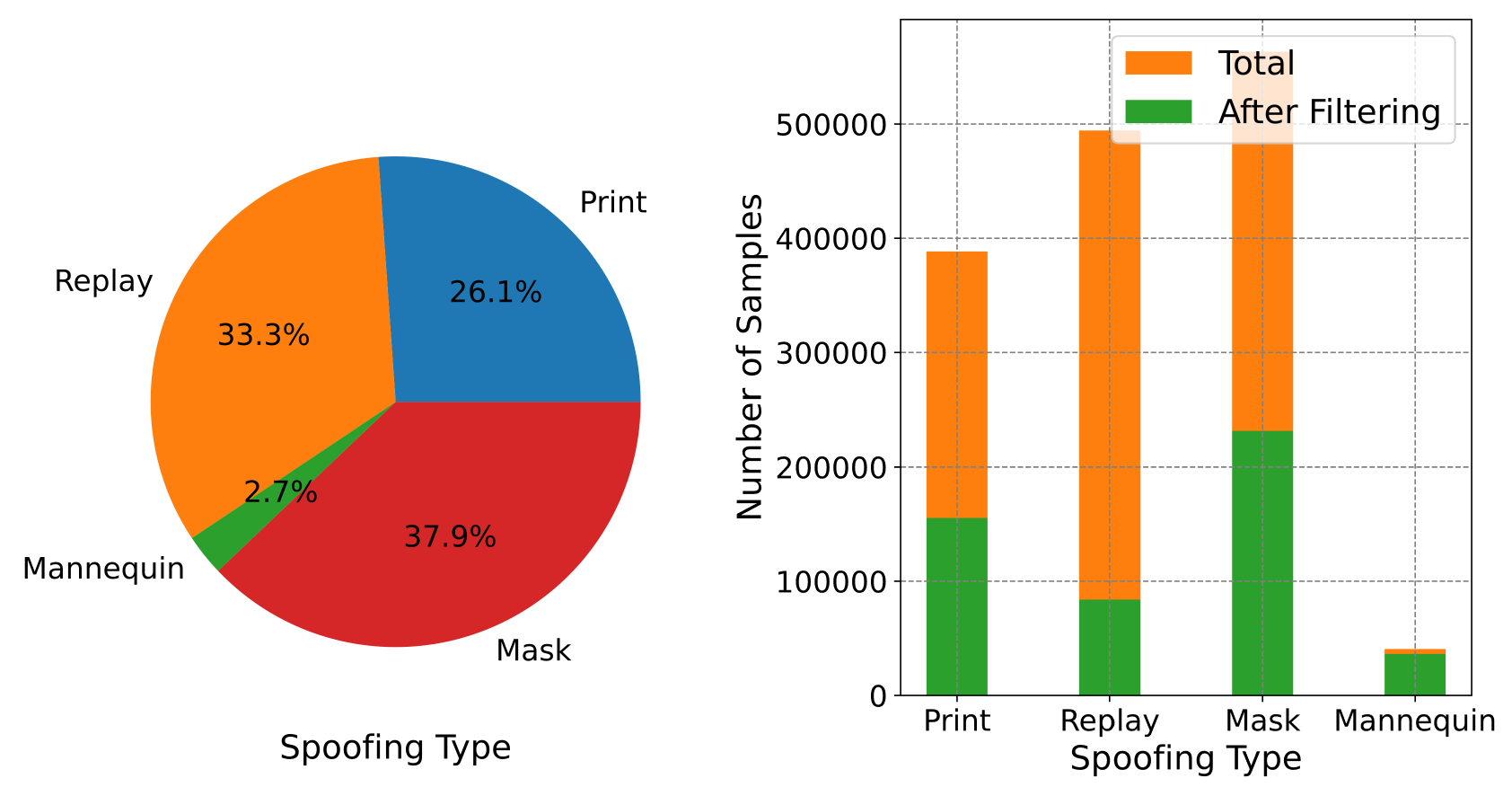}
    \caption{Left: Statistical analysis of the distribution of spoofing types across the 12 FAS datasets. Right: Distribution of sample numbers for different spoofing types before and after filtering.}
    \label{statis}
\end{figure}

Figure \ref{statis} (Right) illustrates the distribution of sample numbers for different spoofing types before and after the filtering process. It is observed that a mere fraction of the initial captions contained the pertinent spoof-aware keywords, highlighting the limited capacity of general captioners to precisely detect spoof-related cues.

\begin{figure*}
    \centering
    \includegraphics[width=40pc]{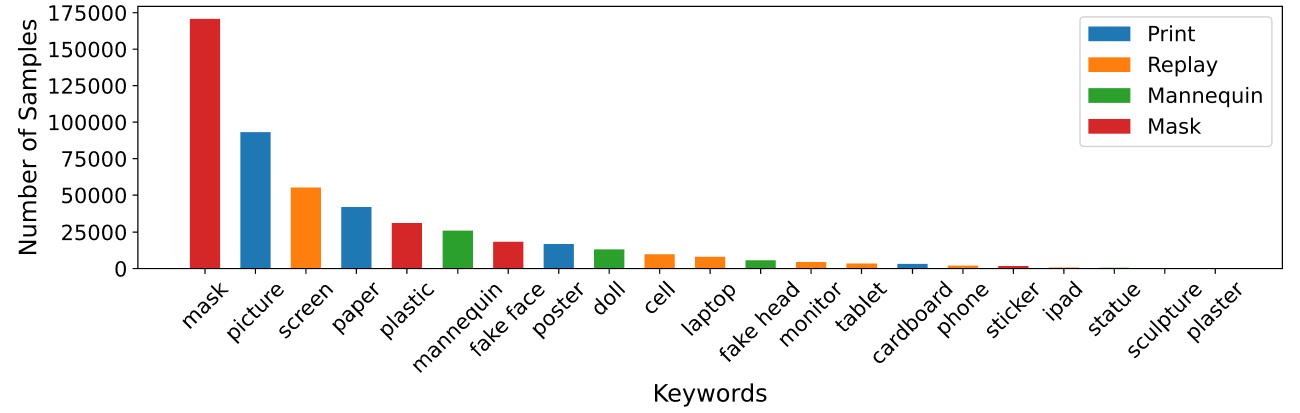}
    \caption{Distribution of the number of samples for different keywords, which are used to filter spoof-aware captions in SCF strategy.}
    \label{keyword}
\end{figure*}

In Figure \ref{keyword}, we present an analysis of the sample filtering process, detailing the number of samples filtered by each specific keyword.  A significant observation is the large number of samples filtered by the keyword \textit{``mask"}, which is consistent with the conclusion in Figure 1 (Right). This observation indicates that the Multimodal Large Language Model (MLLM) is more adept at identifying cues with high-level semantic features, such as visible masks. Conversely, it exhibits a relative weakness in detecting low-level texture features, often subtler and requiring more nuanced analysis, such as moiré patterns from screens and blur associated with paper. This discrepancy underscores the necessity for a more sophisticated approach in feature extraction and integration, such as the Globally Aware Connector (GAC) we introduced, to ensure a comprehensive representation of both high-level and low-level visual cues.

\begin{figure}
    \centering
    \includegraphics[width=20pc]{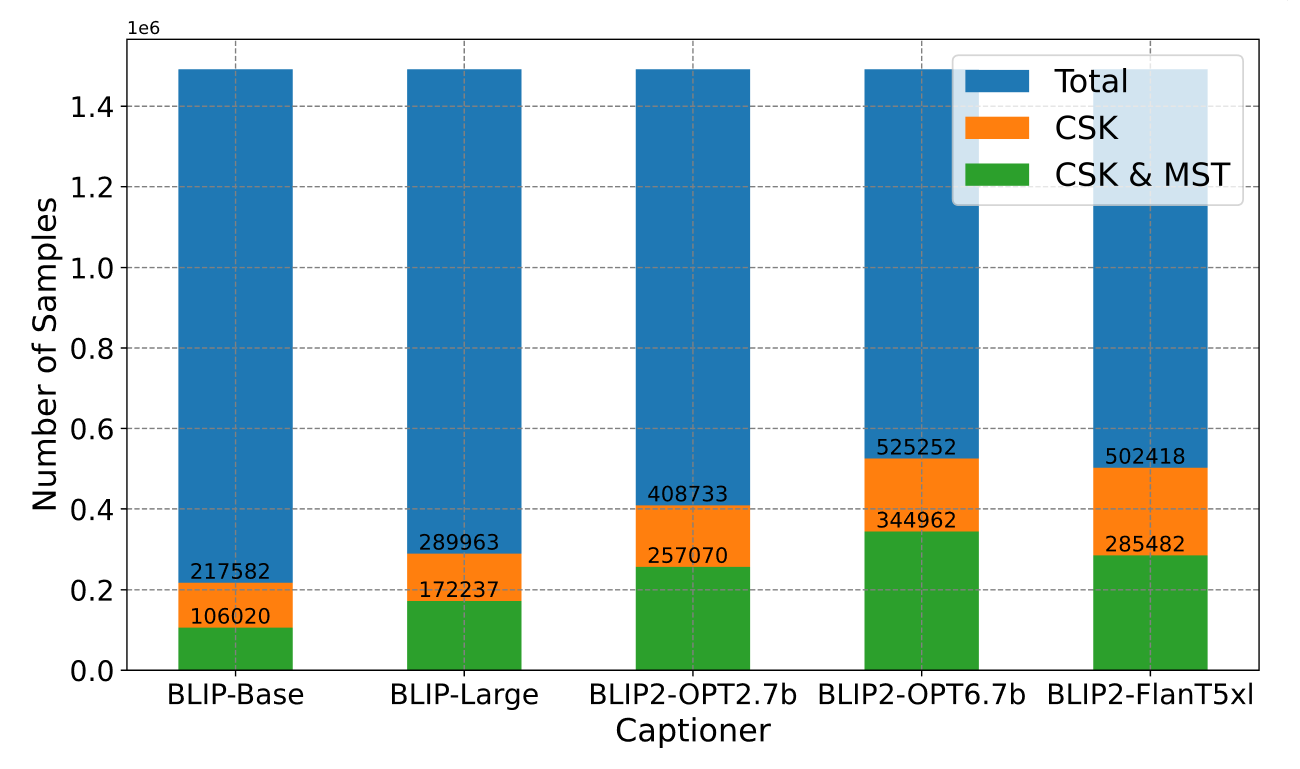}
    \caption{Analysis of the effect of using different captioners in the SCF strategy, where ``CSK" means ``Contain Spoof-aware Keywords" and ``MST" means ``Match Spoofing Types".}
    \label{captioner}
\end{figure}

\begin{table*}[t]
\small
    \centering
    \begin{subtable}[t]{0.23\linewidth}
    \caption{\textbf{Average Over 11 Datasets}}
    \scalebox{0.90}{
    \begin{tabular}{p{14mm}<{\centering}p{10mm}<{\centering}p{10mm}<{\centering}}
    \toprule
    Methods & HTER(\%) & AUC(\%) \\
    \midrule
    w/o SCF & 14.47 & 90.41  \\
    w/o GAC & 14.84 & 90.78  \\
    w/o L-LM & 12.06 & 92.81  \\
    Ours & 11.30 & 93.71 \\
    \bottomrule
    \end{tabular}}
    \end{subtable}
    ~
    \centering
    \begin{subtable}[t]{0.23\linewidth}
    \caption{\textbf{CASIA-MFSD}}
    \scalebox{0.90}{
    \begin{tabular}{p{14mm}<{\centering}p{10mm}<{\centering}p{10mm}<{\centering}}
    \toprule
    Methods & HTER(\%) & AUC(\%) \\
    \midrule
    w/o SCF & 1.15 & 99.88  \\
    w/o GAC & 1.78 & 99.62  \\
    w/o L-LM & 0.93 & 99.93  \\
    Ours & 1.11 & 99.88 \\
    \bottomrule
    \end{tabular}}
    \end{subtable}
    ~
    \centering
    \begin{subtable}[t]{0.23\linewidth}
    \caption{\textbf{CASIA-SURF-3DMask}}
    \scalebox{0.90}{
    \begin{tabular}{p{14mm}<{\centering}p{10mm}<{\centering}p{10mm}<{\centering}}
    \toprule
    Methods & HTER(\%) & AUC(\%) \\
    \midrule
    w/o SCF & 7.80 & 96.85  \\
    w/o GAC & 8.21 & 96.61  \\
    w/o L-LM & 4.72 & 98.70  \\
    Ours & 6.18 & 98.40 \\
    \bottomrule
    \end{tabular}}
    \end{subtable}
    ~
    \centering
    \begin{subtable}[t]{0.23\linewidth}
    \caption{\textbf{HKBU-MARs-V1+}}
    \scalebox{0.90}{
    \begin{tabular}{p{14mm}<{\centering}p{10mm}<{\centering}p{10mm}<{\centering}}
    \toprule
    Methods & HTER(\%) & AUC(\%) \\
    \midrule
    w/o SCF & 34.82 & 69.66  \\
    w/o GAC & 24.65 & 81.81  \\
    w/o L-LM & 27.85 & 79.08  \\
    Ours & 18.64 & 88.77 \\
    \bottomrule
    \end{tabular}}
    \end{subtable}
    ~
    \centering
    \begin{subtable}[t]{0.23\linewidth}
    \caption{\textbf{HiFiMask}}
    \scalebox{0.90}{
    \begin{tabular}{p{14mm}<{\centering}p{10mm}<{\centering}p{10mm}<{\centering}}
    \toprule
    Methods & HTER(\%) & AUC(\%) \\
    \midrule
    w/o SCF & 31.90 & 72.51  \\
    w/o GAC & 27.64 & 79.10  \\
    w/o L-LM & 26.89 & 78.10  \\
    Ours & 28.23 & 77.17 \\
    \bottomrule
    \end{tabular}}
    \end{subtable}
    ~
    \centering
    \begin{subtable}[t]{0.23\linewidth}
    \caption{\textbf{MSU-MFSD}}
    \scalebox{0.90}{
    \begin{tabular}{p{14mm}<{\centering}p{10mm}<{\centering}p{10mm}<{\centering}}
    \toprule
    Methods & HTER(\%) & AUC(\%) \\
    \midrule
    w/o SCF & 2.70 & 99.46  \\
    w/o GAC & 9.44 & 96.01  \\
    w/o L-LM & 3.41 & 99.55  \\
    Ours & 5.63 & 98.73 \\
    \bottomrule
    \end{tabular}}
    \end{subtable}
    ~
    \centering
    \begin{subtable}[t]{0.23\linewidth}
    \caption{\textbf{OULU-NPU}}
    \scalebox{0.90}{
    \begin{tabular}{p{14mm}<{\centering}p{10mm}<{\centering}p{10mm}<{\centering}}
    \toprule
    Methods & HTER(\%) & AUC(\%) \\
    \midrule
    w/o SCF & 15.17 & 92.42  \\
    w/o GAC & 20.05 & 86.06  \\
    w/o L-LM & 16.57 & 90.80  \\
    Ours & 14.86 & 91.68 \\
    \bottomrule
    \end{tabular}}
    \end{subtable}
    ~
    \centering
    \begin{subtable}[t]{0.23\linewidth}
    \caption{\textbf{REPLAY-ATTACK}}
    \scalebox{0.90}{
    \begin{tabular}{p{14mm}<{\centering}p{10mm}<{\centering}p{10mm}<{\centering}}
    \toprule
    Methods & HTER(\%) & AUC(\%) \\
    \midrule
    w/o SCF & 9.43 & 93.82  \\
    w/o GAC & 15.98 & 90.06  \\
    w/o L-LM & 8.77 & 96.00  \\
    Ours & 9.15 & 95.12 \\
    \bottomrule
    \end{tabular}}
    \end{subtable}
    ~
    \centering
    \begin{subtable}[t]{0.23\linewidth}
    \caption{\textbf{Rose-Youtu}}
    \scalebox{0.90}{
    \begin{tabular}{p{14mm}<{\centering}p{10mm}<{\centering}p{10mm}<{\centering}}
    \toprule
    Methods & HTER(\%) & AUC(\%) \\
    \midrule
    w/o SCF & 7.03 & 98.30  \\
    w/o GAC & 7.69 & 97.32  \\
    w/o L-LM & 5.36 & 98.65  \\
    Ours & 5.52 & 98.48 \\
    \bottomrule
    \end{tabular}}
    \end{subtable}
    ~
    \centering
    \begin{subtable}[t]{0.23\linewidth}
    \caption{\textbf{SIW}}
    \scalebox{0.90}{
    \begin{tabular}{p{14mm}<{\centering}p{10mm}<{\centering}p{10mm}<{\centering}}
    \toprule
    Methods & HTER(\%) & AUC(\%) \\
    \midrule
    w/o SCF & 7.92 & 96.98  \\
    w/o GAC & 10.48 & 94.24  \\
    w/o L-LM & 5.34 & 97.97  \\
    Ours & 4.02 & 98.34 \\
    \bottomrule
    \end{tabular}}
    \end{subtable}
    ~
    \centering
    \begin{subtable}[t]{0.23\linewidth}
    \caption{\textbf{SIW-M-V2}}
    \scalebox{0.90}{
    \begin{tabular}{p{14mm}<{\centering}p{10mm}<{\centering}p{10mm}<{\centering}}
    \toprule
    Methods & HTER(\%) & AUC(\%) \\
    \midrule
    w/o SCF & 19.62 & 87.11  \\
    w/o GAC & 16.42 & 90.43  \\
    w/o L-LM & 12.81 & 94.05  \\
    Ours & 10.89 & 95.02 \\
    \bottomrule
    \end{tabular}}
    \end{subtable}
    ~
    \centering
    \begin{subtable}[t]{0.23\linewidth}
    \caption{\textbf{WMCA}}
    \scalebox{0.90}{
    \begin{tabular}{p{14mm}<{\centering}p{10mm}<{\centering}p{10mm}<{\centering}}
    \toprule
    Methods & HTER(\%) & AUC(\%) \\
    \midrule
    w/o SCF & 21.62 & 87.49  \\
    w/o GAC & 20.93 & 90.78  \\
    w/o L-LM & 20.05 & 88.09  \\
    Ours & 20.07 & 89.17 \\
    \bottomrule
    \end{tabular}}
    \end{subtable}
    \caption{Detailed experimental results in ablation experiments, including results on 11 target domain datasets. We run each experiment 3 times under different seeds and report the average HTER and AUC.}
    \label{aba}
\end{table*}

\begin{table}[t]
    \small
    \centering
    \scalebox{0.95}{
    \begin{tabular}{p{10mm}<{\centering}p{40mm}<{\centering}p{10mm}<{\centering}p{10mm}<{\centering}}
    \toprule
    Methods & Trainable Params(M) & HTER(\%) & AUC(\%) \\
    \midrule
    ViTAF & 5 & 23.85 & 82.82 \\
    ViT-B & 86 & 23.48 & 82.98 \\
    ViT-L & 303 & 21.08 & 85.61  \\
    FLIP & 170 & 18.73 & 97.90 \\
    Ours & 104 & \textbf{11.30} & \textbf{93.71} \\
    \bottomrule
    \end{tabular}}
    \caption{Analysis of trainable parameters and performance of different methods. The performance is the average HTER and AUC in Protocol 2.}
    \label{params}
\end{table}

\subsection{Captioner}
In Figure \ref{captioner}, we conduct an exploratory analysis of various general captioners, including BLIP-Base \cite{li2022blip}, BLIP-Large \cite{li2022blip}, BLIP2-OPT2.7b \cite{li2023blip}, BLIP2-6.7b \cite{li2023blip} and BLIP2-FlanT5xl \cite{li2023blip}. We statistically evaluate their initial capacity to perceive spoof-related cues within the framework of our SCF strategy. The evaluation results indicate that the BLIP2-6.7b outperforms the other captioners, demonstrating the highest retention rate of samples after passing through two critical filtering steps: (1) Contain Spoof-aware Keywords (CSK), and (2) Match Spoofing Types (MSK). Given these results, BLIP2-6.7b is selected as our general captioner due to its exceptional ability to identify and retain samples with spoof-related cues. 

Subsequently, we proceed to fine-tune this selected captioner using the filtered subset of 0.3 million fake samples. The fine-tuning process leverages the official codebase in \cite{li2023blip} and adheres to the default configuration for fine-tuning. The model undergoes training for a total of 20 epochs to obtain the spoof-aware captioner.

\section{Ablation Study}
Table \ref{params} provides an analysis of trainable parameters and performance across different methods. Although the integration of the Large Language Model (LLM) has significantly increased the overall parameter volume, we have mitigated this increase by freezing the visual encoder and LLM. Consequently, the volume of trainable parameters, particularly the connector, is comparable to that of other methods. The more significant advantage of our approach, however, lies in its enhanced generalization ability. Despite a potential increase in parameter count, the strategic integration of the LLM and the connector's optimization yields a model that is substantially more adept at generalizing across various domains, which is a critical objective in our research.

Table \ref{aba} presents a comprehensive breakdown of experimental results for each target domain in our ablation experiment, which aims to explore the impact of different components, including Spoof-aware Captioning and Filtering (SCF), Globally Aware Connector (GAC) and Lopsided Language Model (L-LM) loss function. A significant observation is that the variant of our method devoid of the GAC component exhibits a markedly inferior performance on three specific datasets: MSU-MFSD, OULU-NPU, and REPLAY-ATTACK. These datasets are characterized by containing solely print and replay attacks. Upon examination of Figures \ref{vis1} and \ref{vis2}, it becomes evident that the images within these datasets lack conspicuous spoof-related semantic features. Instead, the distinguishing attack features are primarily subtle, low-level textural cues, such as moiré patterns and blurriness. The experimental results validate the GAC's indispensable contribution to the overall performance, especially in discerning nuanced, low-level cues that are vital for accurate spoof detection.

\section{Visualization}
\begin{figure*}
    \centering
    \includegraphics[width=40pc]{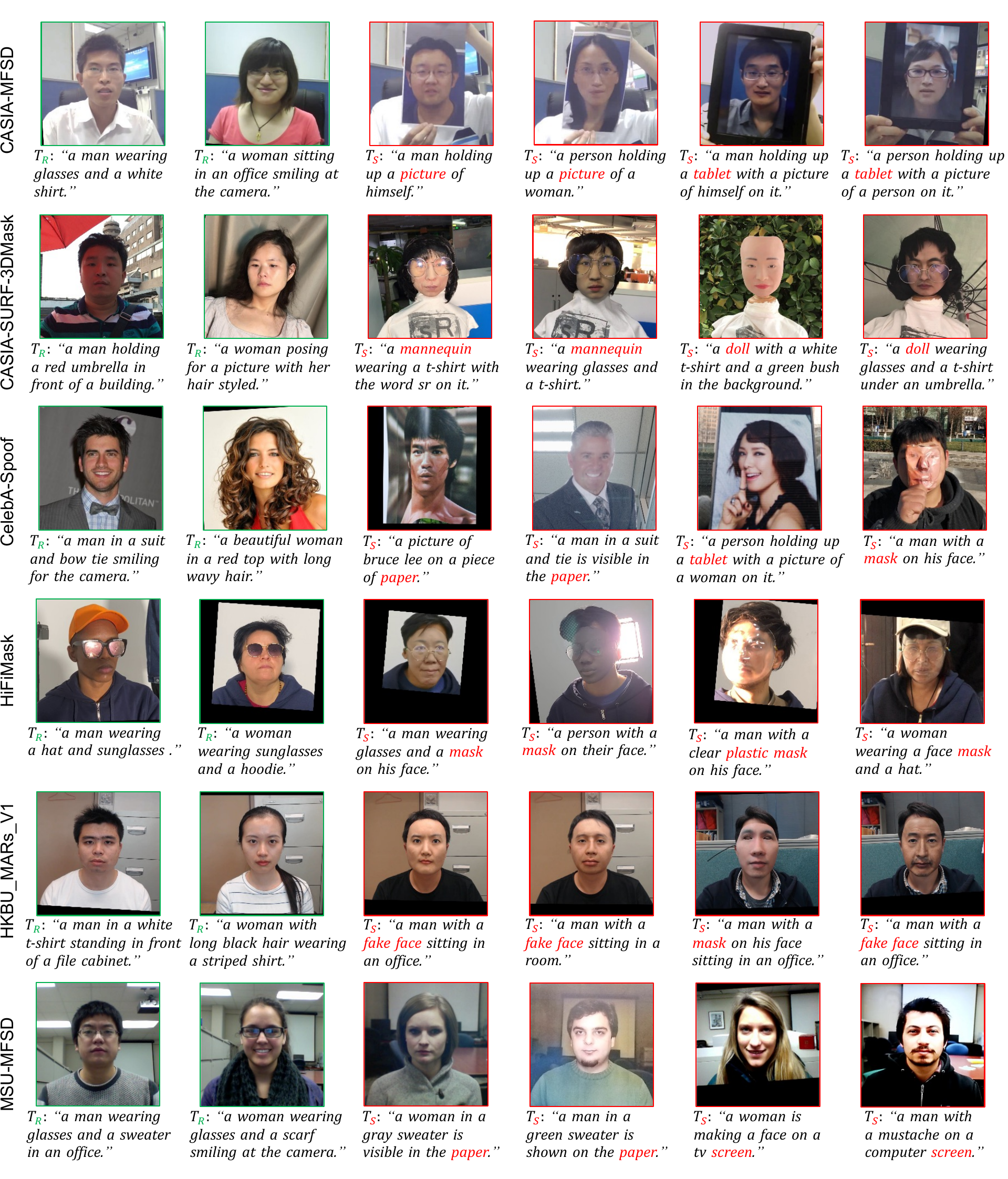}
    \caption{Visualization of some image-caption pairs generated by the SFC strategy on CASIA-MFSD, CASIA-SURF-3DMask, CelebA-Spoof, HiFiMask, HKBU-MARs-V1+ and MSU-MFSD Dataset, The captions $T_{R}$ and $T_{S}$ are generated by general captioner $C_{G}$ and spoof-aware captioner $C_{S}$, respectively. The keywords instrumental in identifying spoof cues are distinctly highlighted in red within the captions}
    \label{vis1}
\end{figure*}

\begin{figure*}
    \centering
    \includegraphics[width=40pc]{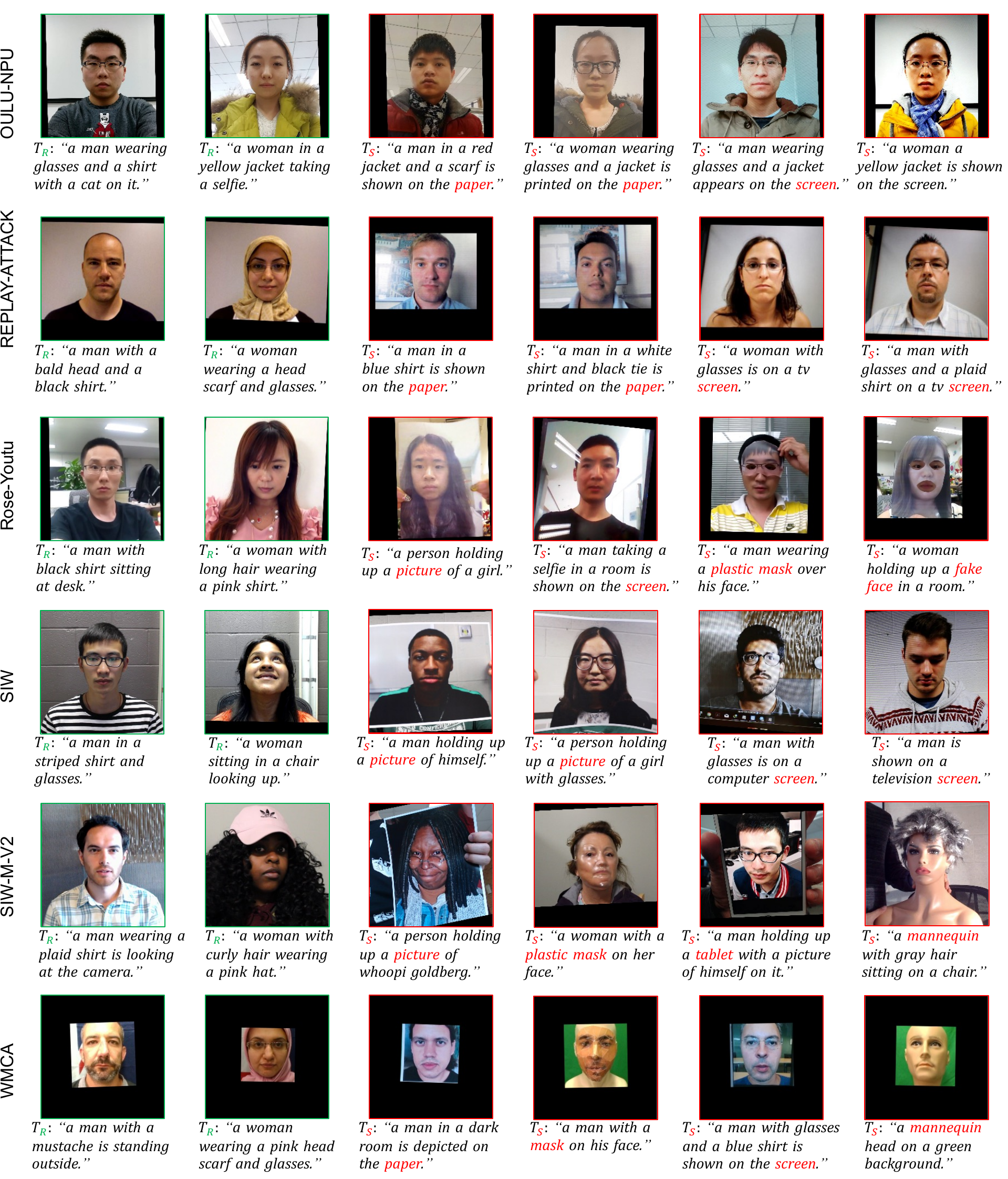}
    \caption{Visualization of some image-caption pairs generated by the SFC strategy on OULU-NPU, REPLAY-ATTACK, Rose-Youtu, SIW, SIW-M-V2 and WMCA Dataset, The captions $T_{R}$ and $T_{S}$ are generated by general captioner $C_{G}$ and spoof-aware captioner $C_{S}$, respectively. The keywords instrumental in identifying spoof cues are distinctly highlighted in red within the captions.}
    \label{vis2}
\end{figure*}

Figure \ref{vis1} and Figure \ref{vis2} show more image-caption pairs generated by the SFC strategy, including real samples (green boxes) and fake samples (red boxes) from 12 datasets. Captions $T_{R}$ generated by the general captioner $C_{G}$ offer generic descriptions, capturing general attributes without bias toward specific features. In contrast, captions $T_{S}$ crafted by the spoof-aware captioner $C_{S}$ reveal the underlying spoofing tactics by emphasizing keywords such as \textit{``paper"}, \textit{``screen"}, and \textit{``mask"}. In addition, captions $T_{S}$ can also reveal spoofing actions, such as \textit{``holding up"}.

\end{document}